\documentclass[10pt,twocolumn,letterpaper]{article}

\usepackage[pagenumbers]{cvpr}

\newcommand{\inlineheading}[1]{\vspace{1pt}\noindent\textbf{{#1}.}}

\definecolor{iccvblue}{rgb}{0.21,0.49,0.74}
\usepackage[pagebackref,breaklinks,colorlinks,allcolors=iccvblue]{hyperref}
\usepackage{float}
\usepackage{multirow}
\usepackage{microtype}

\def\paperID{272}
\def\confName{3DV\xspace}
\def\confYear{2026\xspace}

\newcommand{\method}{Fillerbuster\xspace}
\title{\method: Unified Generative Scene Completion Model for Casual Captures}

\author{%
	Ethan Weber\textsuperscript{1,2}\qquad%
	Norman Müller\textsuperscript{1}\qquad%
	Yash Kant\textsuperscript{1,3}\qquad%
	Vasu Agrawal\textsuperscript{1}%
	\\[0.15em]%
	Michael Zollhöfer\textsuperscript{1}\qquad%
	Angjoo Kanazawa\textsuperscript{2}\qquad%
	Christian Richardt\textsuperscript{1}%
	\\[0.5em]%
	\textsuperscript{1}Meta Reality Labs\quad%
	\textsuperscript{2}UC Berkeley\quad%
	\textsuperscript{3}University of Toronto%
}

\begin{document}
\begin{figure}
\twocolumn[{%
\renewcommand\twocolumn[1][]{#1}
\maketitle
\centering
\vspace{-2em}
\includegraphics[width=\textwidth]{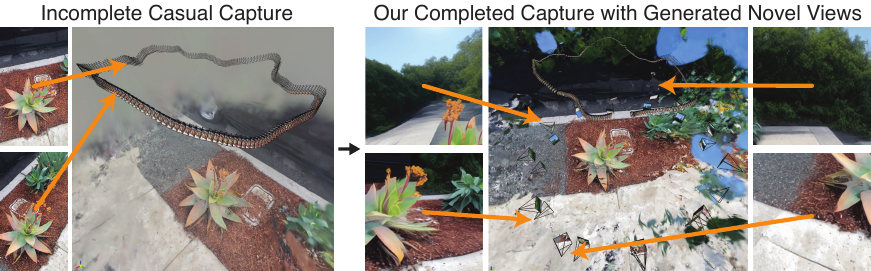}
\caption{\label{fig:teaser}%
    \textbf{Completing casual captures.}
    \method takes an incomplete casual capture with many images (left) and conditions on these to create consistent novel views (right). The original images and the new ones enable novel-view synthesis that is more complete than vanilla Gaussian Splatting trained on only the incomplete casual capture. Project page: \url{https://ethanweber.me/fillerbuster/}.
}
}]
\end{figure}

\begin{abstract}%
We present \method,
a unified model that completes unknown regions of a 3D scene with a multi-view latent diffusion transformer.
Casual captures are often sparse and miss surrounding content behind objects or above the scene.
Existing methods are not suitable for this challenge as they focus on making known pixels look good with sparse-view priors, or on creating missing sides of objects from just one or two photos.
In reality, we often have hundreds of input frames and want to complete areas that are missing and unobserved from the input frames.
Our solution is to train a generative model that can consume a large context of input frames while generating unknown target views and recovering image poses when camera parameters are unknown.
We show results where we complete partial captures on two existing datasets.
We also present an uncalibrated scene completion task where our unified model predicts both poses and creates new content.
We open-source our framework for integration into popular reconstruction platforms like Nerfstudio or Gsplat.
We present a flexible, unified inpainting framework to predict many images and poses together, where all inputs are jointly inpainted, and it could be extended to predict more modalities such as depth.
\end{abstract}
\vspace{-2em}

\section{Introduction}
\label{sec:introduction}

Photogrammetry has been around for decades \cite{Sturm2011} but only recently has become mainstream with novel-view synthesis techniques becoming high fidelity, such as NeRF \cite{mildenhall2021nerf} and Gaussian Splatting \cite{kerbl20233d}.
Widely used apps like Polycam \cite{polycam} or Flythroughs \cite{Flythroughs} mean that everyday people can go out and easily capture content.
Many such captures are done casually, which means the data is collected rather quickly and may miss large portions of the scene where the camera never looked.
Sometimes, the capture is just a handful of sparse photos, which makes obtaining camera poses challenging.

Reconstructing casually captured scenes is challenging because there is missing content to complete and it is not predictable where the missing content will be from capture to capture.
In contrast, the object-centric setting is much simpler as one can assume a canonical coordinate frame and sample missing views looking inward on a sphere.
Instead, we highlight the challenges of scenes and focus on this more general setting, where the input camera poses can be incredibly diverse.
Our goal is to fill in the missing information to enable an immersive view of the scene that feels complete, and where the rendered content can go beyond what is seen in training images, as illustrated in \cref{fig:teaser}.
To address this problem setting, we propose \method, a unified model that completes scenes by jointly denoising images and camera raymaps from many input frames (and predicts poses when unknown).
This content may be casual videos, where a user quickly scans their phone through a scene, or this may be a sparse set of photos with unknown poses, e.g. from a vacation.
Given this data capture as input, our unified model jointly completes the unobserved content and recovers poses.

To improve casual captures, our key insight is to jointly model the image and camera distribution of existing casual captures by using a multi-view aware diffusion model.
Our approach is made possible by the large influx of captured data being recorded and uploaded online.

\begin{figure}[t]
\centering
\includegraphics[width=\linewidth]{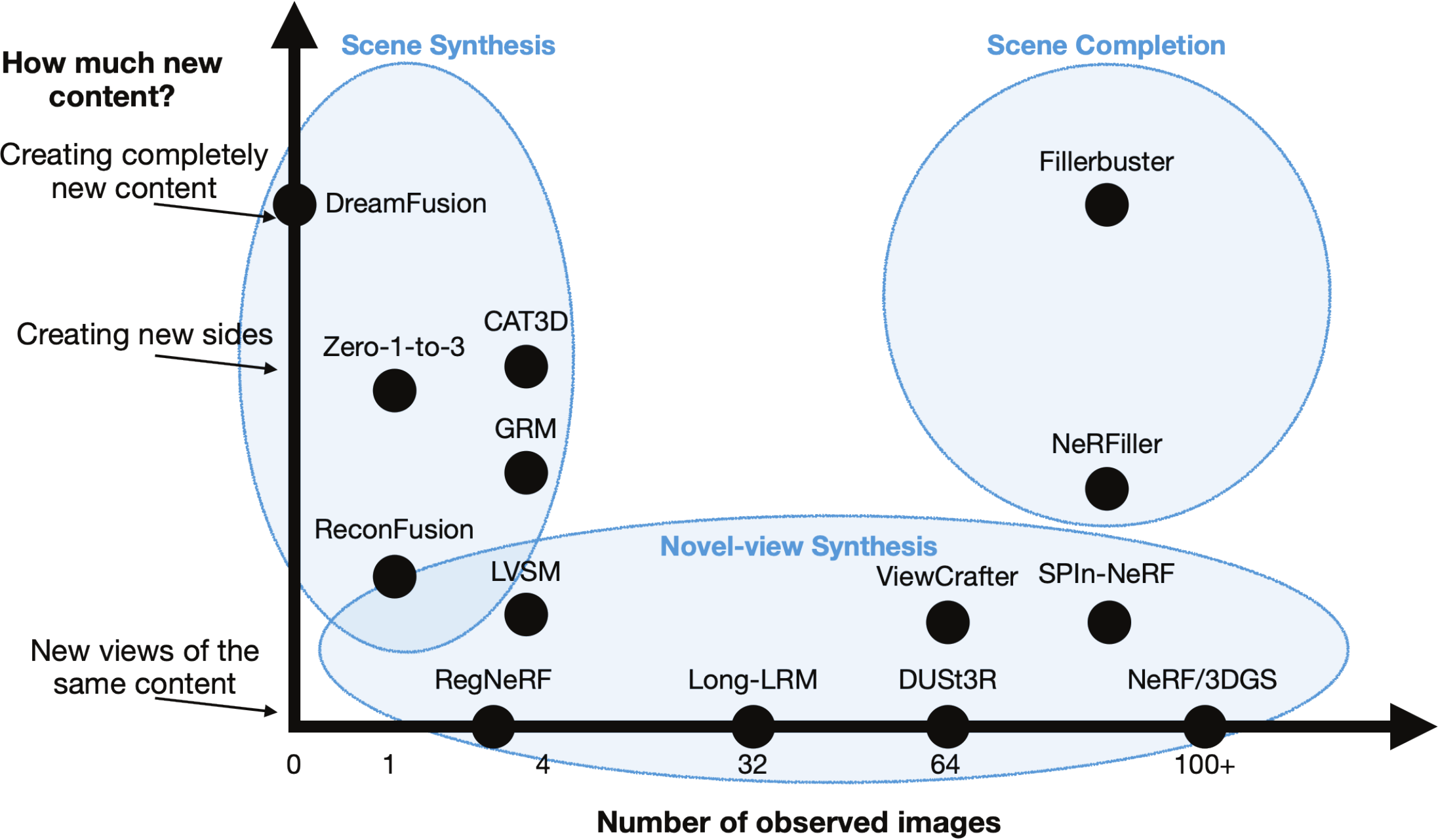}
\caption{\label{fig:problem_setting}%
    \textbf{Problem setting.}
    We illustrate our problem setting with respect to a non-exhaustive set of related work.
    Many works focus on scene synthesis (left) where one generates data from text or from a single image.
    Similarly many tackle novel-view synthesis (bottom) to synthesize new views of the input image content.
    Fewer works focus on scene completion where the task is to complete missing content in captures (top right).
}
\vspace{-20px}
\end{figure}

We design our model to handle a large and variable number of input and output frames.
Unlike typical generative NVS methods that are autoregressive, our model performs non-autoregressive joint denoising over long sequences to generate many views at once.
More specifically, our problem setting is very different from the common settings of
(1) generation from text only (no images) \cite{HoelleCOJN2023,WangLWBLSZ2023,ShiWYLLY2024}
(2) from just one image \cite{liu2023zero,SargeLSHYZCLFSW2024}, or
(3) from two images with the goal of interpolating between them \cite{jin2024lvsm,yu2024viewcrafter}.
We present an overview of related work in \cref{fig:problem_setting}.
Concurrent work also lacks the flexibility of our unified model.
Our model can take in many images and camera poses, e.g. 50 images, and the user can specify which content is known, which is unknown, and which should be completed.
The model can also take in partially complete images, unlike the current paradigm of assuming the input images are fully intact and known \cite{liu2023zero,gao2024cat3d}.
Fillerbuster could flexibly be extended with our open-source release to inpaint more modalities such as depth, gaussians, or semantics using our unified framework.

We demonstrate our problem setting and the usefulness of our \method model on multiple tasks.
First, we show our model can complete casual captures by hallucinating large unknown regions.
Second, we introduce the task of ``uncalibrated scene completion", where the goal is to recover both the image poses and completed novel views.
Notably, we perform both tasks with our unified model.
Third, we show the multi-view inpainting task on the NeRFiller dataset \cite{weber2024nerfiller}, where we surpass prior work in quality and consistency.
Finally, we present an ablation of our modeling decisions and show our model's ability to gracefully handle different numbers of input images. All code will be released.

\begin{figure*}[t]
\centering
\includegraphics[width=\textwidth]{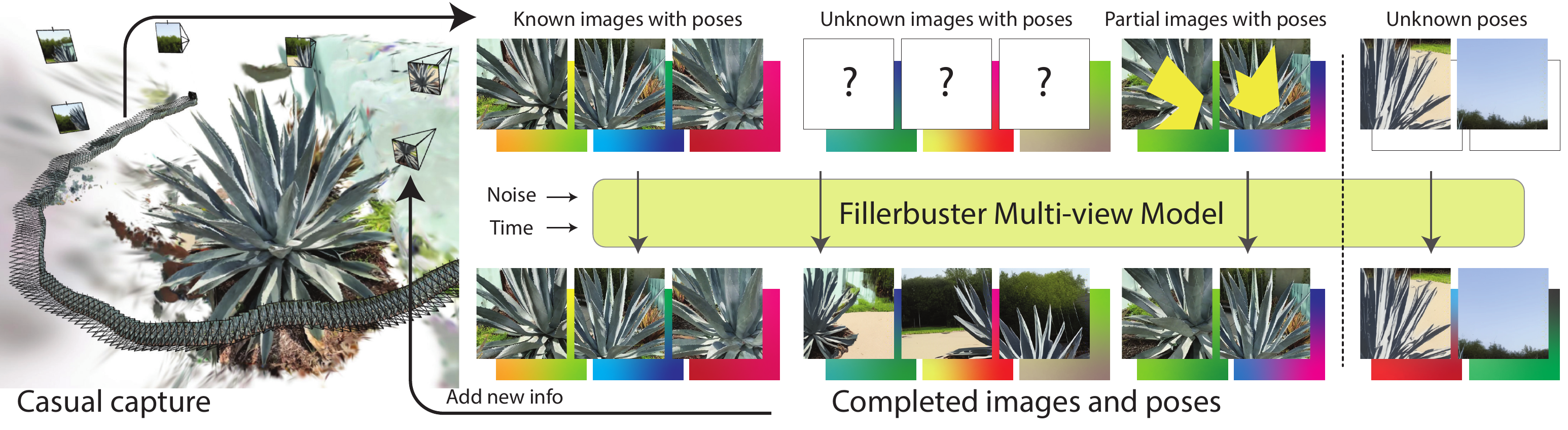}
\caption{\label{fig:model_architecture}%
    \textbf{Model overview.}
    \method is trained on a large collection of multi-view images and poses (top and bottom of stacked images, respectively), which makes it useful for completing casual captures at inference time.
    More specifically, we are interested in four primary uses of the model:
    (1) conditioning on known images which have pose,
    (2) predicting new views where poses are provided,
    (3) predicting partial images where some pixels are known, or
    (4) recovering the camera poses when unknown.
    Our model is a latent DiT trained to jointly model images and poses for any mixture of the input.
    In practice, our poses are 6-channel raymaps encoding ray origins and directions.
    \looseness-1
    \vspace{-1.7em}
}
\end{figure*}

\section{Related work}
\label{sec:related_work}
\vspace{-0.5em}

\inlineheading{Few-view reconstruction}
Our problem setting is not few-view, but we highlight the differences here.
These methods focus on deleting reconstruction artifacts \cite{warburg2023nerfbusters,sabour2024spotlesssplats,goli2024bayes}, using sparse-view losses \cite{niemeyer2022regnerf}, leveraging depth and normal priors \cite{TurkuRMSRK2025}, or using generative models \cite{wu2024reconfusion,LiuCKTT2024,KantSVGRTG2023,wynn2023diffusionerf}, to complete sparse captures.
In contrast, our goal is scene completion from realistic casual captures with many input images: generatively completing unobserved scene content beyond the captured views (optionally recovering poses), which is not directly addressed by sparse-view reconstruction methods.

\inlineheading{Multi-view generative models}
Many single-view to single new-view models exist where an input image is known and a target image is unknown \cite{liu2023zero,SargeLSHYZCLFSW2024,seo2024genwarp,tewari2023diffusion,wang2023imagedream}.
A few methods have increased the input context to multiple input images but still generate just one output view \cite{wu2024reconfusion,jin2024lvsm}.
Even fewer methods have increased both the number of inputs and the number of outputs.
CAT3D \cite{gao2024cat3d} uses 1 or 3 input images and generates 7 or 5 images, but never goes beyond a total sequence size of 8 (described in their supplemental material).
It also remains closed-source, which limits its impact.
Our model, in contrast, supports a larger and flexible number of input and output images, and will be open-sourced.
Concurrently, multi-view diffusion systems have explored related directions \cite{li2025nvcomposer,lu2025matrix3d,stablevirtualcamera2025}, but we target a distinct goal of scene completion with large-sequence conditioning and joint image–pose denoising to generate many views for downstream 3D reconstruction, which complements these works.
We emphasize the importance of large sequence sizes to fit the entire casual capture in context, which helps keep generative completion consistent with the observations and more stable for downstream reconstruction.
Some video models have been fine-tuned for camera control \cite{he2024cameractrl,wang2024motionctrl,VanHWOSLTDZV2024} or use geometry conditioning \cite{yu2024viewcrafter,liu2024reconx,muller2024multidiff,MVDiffusion}, but these models generate smooth temporal videos, so can neither condition on the entire capture nor generate the many well-distributed views typical for the 3D reconstruction setting. \cite{liu20243dgs,wu2025difix3d+} clean up 3DGS artifacts in captures using diffusion priors.

\inlineheading{LRMs conditioned on cameras}
Large reconstruction models (LRMs) that predict 3D have become popular to directly predict Gaussians \cite{xu2024grm,ZiwenTZBLHFX2024,ren2024l4gm}.
Most methods assume camera poses as input, which may come from traditional methods like COLMAP \cite{schonberger2016structure,pan2024global} or data-driven methods \cite{wang2024dust3r,leroy2024grounding}.
LRMs are excellent at predicting pixel-aligned geometry but cannot inpaint unobserved areas of the scene.
Furthermore, they rely on camera poses, which may be unknown in casual captures.
We present a unified model for both tasks, such that when the camera is unknown, we can perform the ``uncalibrated scene completion" task of making a camera fly-through of the scene from a set of sparse unposed photos.
We model our camera pose prediction inspired by other data-driven approaches \cite{zhang2024cameras}, but use a raymap latent space, with ray origins and directions instead of Plücker coordinates.

\inlineheading{3D inpainting}
Current 3D inpainting methods such as SPIn-NeRF \cite{mirzaei2023spin} or NeRFiller \cite{weber2024nerfiller} rely on using 2D inpainting models within the NeRF 3D reconstruction framework to complete scenes \cite{ShihMHCCK2024}.
Most methods \cite{CaoYFWX2024,ChenCSXZ2024,LinKHLMKYT2024,ChenLP2024,mirzaei2023reference} focus on the SPIn-NeRF dataset, which is forward-facing and has much less camera motion than a typical casual capture.
NeRFiller has more challenging camera movement, so we consider this dataset for experiments.
However, none of these methods are conditioned on camera views when inpainting.
This makes it impossible for them to complete scenes with large unknown content.
This is because some of the generated views will be completely unknown, and without having context of the existing scene, it is unclear how to fill in the image.
In contrast, our approach is camera-pose conditioned.

\begin{figure*}[t]
\centering
\includegraphics[width=\linewidth]{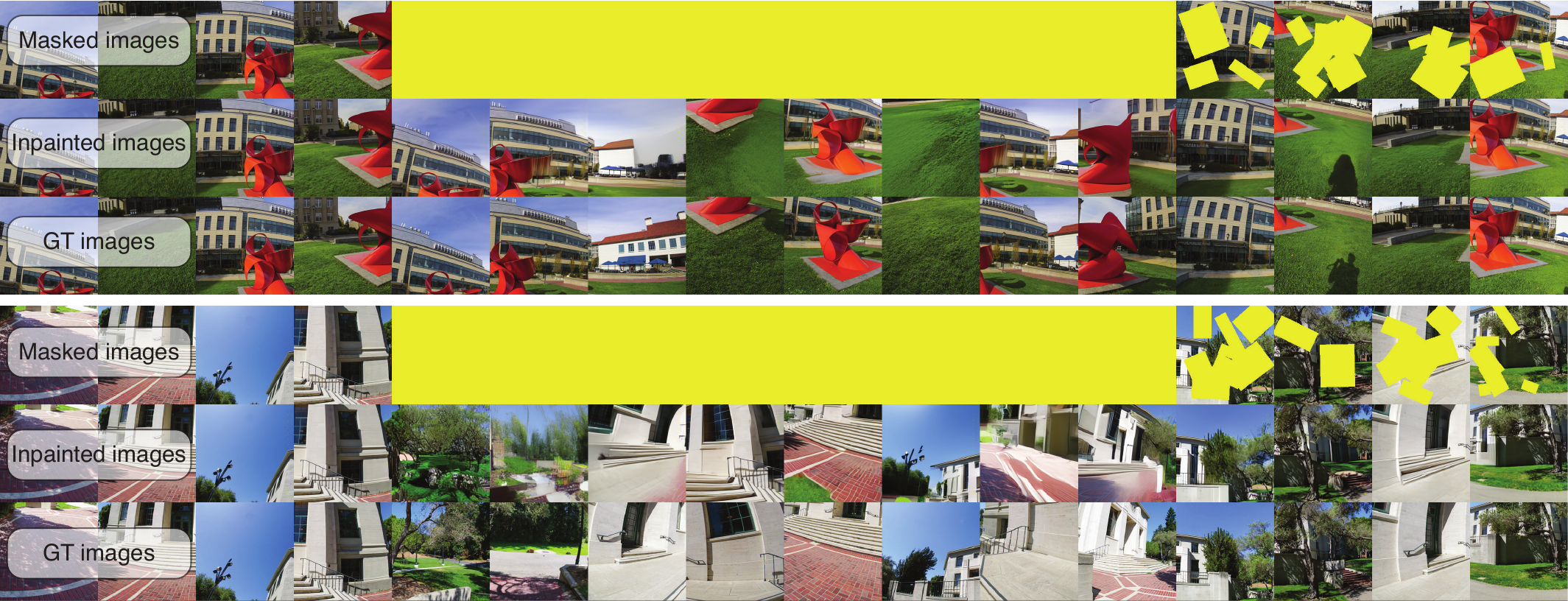}
\vspace{-15px}
\caption{\label{fig:model_samples}%
    \textbf{Model samples.}
    Here we show generations from our model.
    For this setting, we provide pose input for all images.
    The ``Masked'' rows indicate which pixels are known, with yellow indicating unknown regions. This is the model conditioning signal.
    The ``Inpainted'' rows show the inpainted images after passing the entire sequence of size 16 (top rows) into the model for 24 denoising steps.
    The ``GT'' rows show the ground truth, but note that this is not necessarily the only correct solution if the newly generated pixels are unobserved according to the masks.
    Notice that in the top example, the generations are self-consistent but different from the GT, which is entirely plausible.
    \vspace{-1em}
}
\end{figure*}

\vspace{-5px}
\section{Method}
\label{sec:method}
\vspace{-5px}

We first explain our model's details for jointly modeling image completion and poses (\cref{sec:model}), and then explain how to use \method for casual scene completion with our model being helpful for 3D reconstruction (\cref{sec:multi-view-completion}).

\subsection{\method Model}\label{sec:model}

We propose a latent diffusion transformer that denoises multiple input images and calibrated camera poses with masks indicating known and unknown regions.
There are $N$ elements in a sequence with images $I_i \in \mathbb{R}^{H \times W \times 3}$, raymaps $R_i \in \mathbb{R}^{H \times W \times 6}$ with origin and direction per pixel, valid image masks $\mathcal{M}^\text{I}_i \in \mathbb{R}^{H \times W}$, and valid ray masks $\mathcal{M}^\text{R}_i \in \mathbb{R}^{H \times W}$, where 1 indicates known conditioning information and 0 indicates unknown pixels.
Our goal is to predict all images and raymaps given only the known information, i.e., $p(I,R \mid I \odot \mathcal{M}^\text{I}, R \odot \mathcal{M}^\text{R})$.
We use ``sequence'' to refer to multiple images and cameras from the same capture, and ``sequence size'' for how many images are denoised together.

\inlineheading{Model architecture}
The architecture is designed for latent inpainting \cite{rombach2022high}, taking in any combination of known and unknown images and raymaps, and predicting the missing values.
We use a DiT architecture \cite{peebles2023scalable} and train with the flow matching objective \cite{lipman2022flow}.
We train separate VAEs for images and poses encoded as raymaps, where $\mathcal{E}^\text{I}$ denotes the image encoder and $\mathcal{E}^\text{R}$ denotes the raymap encoder.
Both encoders compress the spatial resolution by a factor of 8$\times$ and output a $d$-dimensional representation.
We set $d=16$ for both encoders.
Let $z^\text{I}_{i} = \mathcal{E}^\text{I}(I_{i})$ and $z^\text{R}_{i} = \mathcal{E}^\text{R}(R_{i})$ denote the compressed latent image and raymap, respectively.
Let $\mathcal{D}$ denote a downscaling operation that reduces the spatial resolution by the same factor as the encoders.
We add noise to $z_{i}$ as $\tilde{z}_{i,t} = (1 - t) z + t \epsilon$, then prepare the sequence as
\begin{align}
\begin{split}
s_{i,t} = {} & \tilde{z}^\text{I}_{i,t} \oplus \mathcal{E}^\text{I}(I_{i} \odot \mathcal{M}_{i}^\text{I}) \oplus \mathcal{D}(\mathcal{M}_{i}^\text{I}) \oplus {} \\
 & \tilde{z}^\text{R}_{i,t} \oplus \mathcal{E}^\text{R}(R_{i} \odot \mathcal{M}_{i}^\text{R}) \oplus \mathcal{D}(\mathcal{M}_{i}^\text{R}) \text{,}
\label{eq:sequence_equation}
\end{split}
\end{align}
where $\oplus$ denotes concatenation of the noisy latents, known image and ray latents, and the masks themselves.
The noisy sequence $s_{t} \in \mathbb{R}^{N \times H \times W \times (4d+2)}$ is patchified, and positionally embedded (described later), then passed through the transformer model $\mathcal{F}$ to predict the denoised latent images and raymaps as $\{z^\text{I}, z^\text{R}\} = \mathcal{F}(s)$.
Our VAEs have a convolutional architecture \cite{rombach2022high} and train with KL \cite{kingma2013auto}, adversarial \cite{goodfellow2014generative}, and L1 reconstruction losses.
Our transformer architecture is ``DiT-L/2" \cite{peebles2023scalable} with a latent patch size of 2$\times$2 and 24 layers of multi-head self-attention.
Our model only has 650M parameters – small enough to fit on most GPUs and fast enough to use with open-source 3D reconstruction tools.

\inlineheading{Raymap coordinate convention}
Raymaps comprise per-pixel ray origins and world-space unit directions.
At training time, we randomly choose one camera from our sequence to be at the origin and oriented upright.
We also randomly rotate and rescale the cameras for augmentation, and ensure that origins are always within the cube $[-1, 1]^3$.

\inlineheading{Masking out regions}
During training, we mask out information from the images and/or the raymaps; the task is to predict the denoised sequence from partial noise.
We apply masking at the pixel level before VAE encoding (\cref{eq:sequence_equation}) to enable precise control over which pixels are known or unknown.
We use a mixed masking strategy: some images are known, some are unknown, and some are partially unknown with randomly rotated rectangles, as illustrated in \cref{fig:model_samples}.
We dropout image and raymap masking with a 10\% chance to enable classifier-free guidance \cite{ho2022classifier}.

\inlineheading{Token positional embeddings}
We use two forms of positional embeddings to enable varying sequence lengths to be generated at inference time.
\textit{1) 2D layout embeddings} encode the layout of the image with fixed sinusoidal embeddings.
\textit{2) Index embeddings} are more unique for our setting, where we add an unordered index descriptor to each token coming from the same image.
More specifically, the full sequence $s$ is first patchified and projected into patches $p$.
It is then augmented with positional embeddings as $p' = \psi_\text{2D}(p) + \psi_\text{Idx}(p)$, where $\psi_\text{2D}$ is sinusoidal embeddings to encode the 2D layout of each patch within the image itself \cite{vaswani2017attention}, and $\psi_\text{Idx}$ to encode which index in the sequence the patch is from.
During training, $\psi_\text{Idx}(p)$ randomly samples a frequency for each image and then adds that value to each patch of the same image.
During inference, the frequencies are chosen with uniform spacing and applied in the same way so each image has a unique identifier.
Concretely, for a sequence of length $n$, we set a scalar per-image index $u_k = \tfrac{k}{\max(n-1,1)}$ for $k\in\{0,\ldots,n{-}1\}$ and use standard sinusoidal embeddings $\psi_\text{Idx}(u_k)$ added to all patches belonging to image $k$.
This helps support generating longer sequences at inference time beyond the training lengths, which we show in \cref{sec:model-design}.
We show that this is useful for generating longer sequences beyond the length that we train with.

\inlineheading{Training and inference details}
We train our model \textit{from scratch} on a collection of datasets including
ScanNet++ \cite{yeshwanth2023scannet++}
and a corpus of Shutterstock data including 2D images and 3D asset renderings.
We train our final model on 64 A100 GPUs for approximately a month.
We first train at 256$\times$256 resolution for 1M iterations, and then fine-tune for 100K iterations with resolutions varying from 64$\times$64 to 1024$\times$1024 with sequence lengths between 20 and 2, depending on how many images fit in GPU memory for a given resolution.
See the appendix for additional details.
\looseness-1

\subsection{Multi-View Scene Completion}\label{sec:multi-view-completion}

Here we explain how to use our model to complete scenes.

\begin{figure*}[t]
\centering
\includegraphics[width=\linewidth]{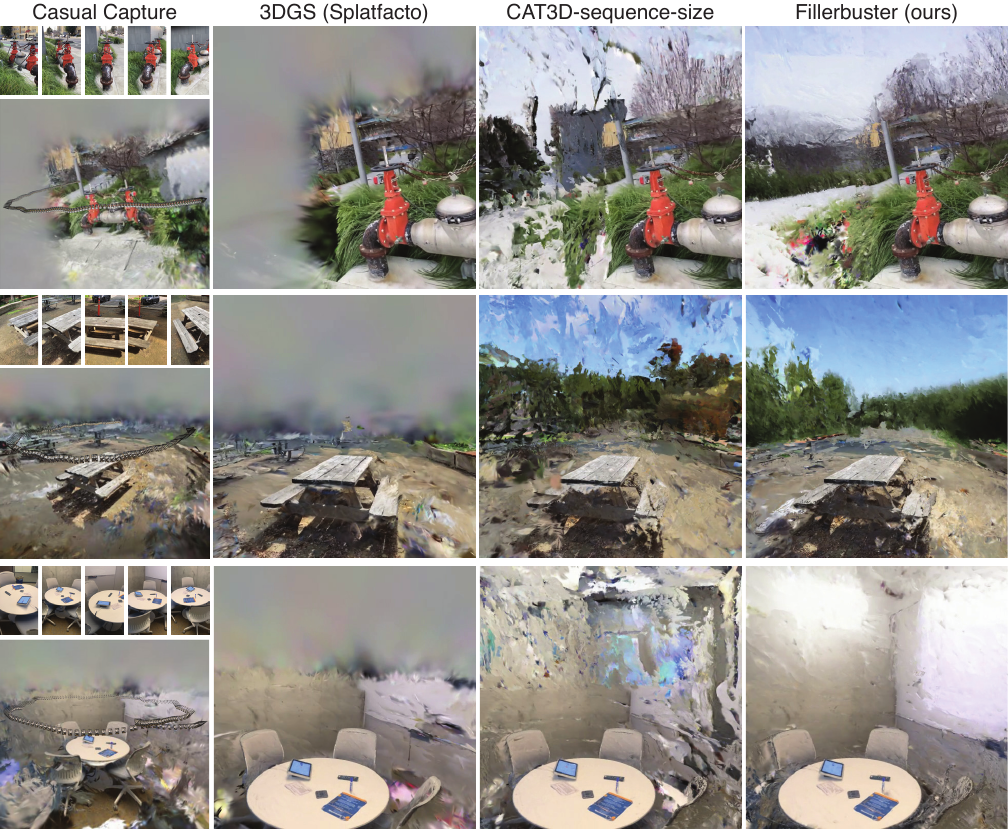}\vspace{-0.5em}
\caption{\label{fig:completing_casual_captures}%
    \textbf{Completing casual captures.}
    Here we demonstrate our ability to complete casual captures from the training splits of the Nerfbusters dataset \cite{warburg2023nerfbusters}.
    On the left, we show the input captures and some representative images.
    3DGS (Splatfacto) cannot add missing details so the capture remains incomplete.
    Our CAT3D baseline conditions on 3 images and generates 6 images at a time, so it cannot produce consistent content.
    \method conditions on 16–40 images to generate 24 novel views and obtains the most consistent results.
}\vspace{-1em}
\end{figure*}

\inlineheading{Variable sequence lengths}
Our index embedding enables changing the sequence length at inference time.
We leverage this property to generate many images at the same time for inpainting incomplete scenes, since NeRF and Gaussian splatting typically require many views to create a scene.

\inlineheading{Multi-view inpainting for scene completion}
We complete scenes by generating novel views and adding them to our existing dataset, then optimizing 3DGS \cite{kerbl20233d}.
We avoid the need for an SDS-like optimization approach~\cite{poole2022dreamfusion,haque2023instruct} because our model can generate many consistent images with a large sequence size.
To complete scenes with large camera movement, we add new views to the scene that look in all directions.
We first inpaint many ($\sim$25) ``anchor'' frames, and then condition on these frames to generate more novel views, as in CAT3D \cite{gao2024cat3d}.
The key difference is that we can handle much larger sequence sizes than CAT3D, which operates on at most 3 images for conditioning.
Furthermore, unlike CAT3D, we can also complete scenes with partial masks.
To complete these scenes, we inpaint the images themselves and update the dataset with the new pixels.

\inlineheading{Normal regularization}
We find that regularizing Gaussian splat geometry towards the end of optimization can help improve results. Specifically, we apply a total-variation smoothness loss on rendered normals \cite{fridovich2022plenoxels}, and we also align our depth-derived surface normals with rendered normals (similar to \citet{verbin2024nerf} but using Gaussians instead of NeRF).
This second loss is $\mathcal{L}_\text{align} = \|\mathrm{sg}(N_\text{r}) - N_\text{d}\|^{2}_{2} + \|N_\text{r} - \mathrm{sg}(N_\text{d})\|^{2}_{2}$, where $N_\text{r}$ are rendered normals from 3D Gaussians, oriented towards the camera, and $N_\text{d}$ are normals derived from rendered depth maps.
We apply our normal regularizations after the initial geometry has taken form, at approximately 10K steps.
The appendix has more details.

\vspace{-7px}
\section{Evaluation}
\vspace{-8px}
We first show our casual scene capture completion results on the Nerfbusters dataset, and then demonstrate the ``uncalibrated scene completion'' task on data captured ourselves.
Next, we show results on the NeRFiller dataset, where we surpass prior work in quality and consistency.
Finally, we evaluate our model design choices.
Note that we choose to use 3D Gaussian splatting \cite{kerbl20233d} for our reconstruction experiments rather than NeRF \cite{mildenhall2021nerf} because 3DGS is fast to train and thus gaining popularity among casual capture users.

\subsection{Completing Casually Captured Scenes}
\label{sec:completing_casual_captures}
\vspace{-5px}

\begin{figure}[t]
\centering
\includegraphics[width=\linewidth]{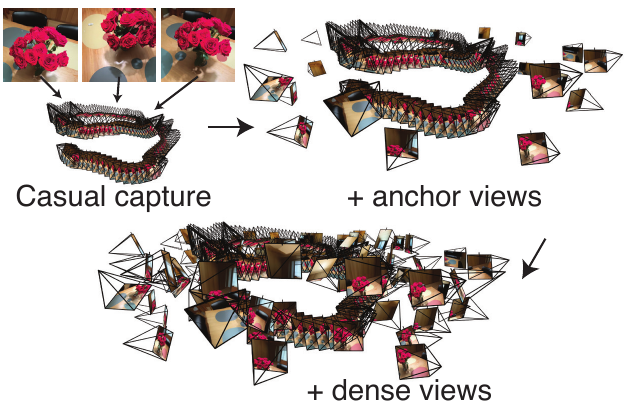}
\caption{\label{fig:camera_sampling}%
    \textbf{Novel-view sampling.}
    We start with a casual capture (top left) and condition on 16 of the images to generate 24 anchor views simultaneously (top right).
    We then condition on the casual capture and anchors to densify views (bottom).
    We repeat the dense stage for multiple rounds to reach $\sim$100 novel views in total.
    \vspace{-2em}
}
\end{figure}

\inlineheading{Setting}
Here we show results for completing casually captured scenes.
We choose the Nerfbusters dataset \cite{warburg2023nerfbusters} for this setting because it mimics the casual captures of an inexperienced user.
Our goal is to take these partial captures and to complete them – either by completing geometry or adding context to the capture.
We compare the following methods:
(1) \textit{3DGS} (Splatfacto \cite{tancik2023nerfstudio}, which uses the gsplat library~\cite{ye2024gsplat}, with no inpainting),
(2) \textit{NeRFiller} \cite{weber2024nerfiller} (NeRFiller inpainting, which we note is not suitable for this setting where the new views do not have partial masks),
(3) \textit{CAT3D-sequence-size} (ours with CAT3D-sized conditioning, where we condition on 3 images and generate 6 images a time, further described in the appendix), and
(4) \textit{\method} (our complete method, where we condition on 16 views and generate 24 images at a time).
We perform multiple rounds of inpainting to reach $\sim$100 new views that are added to the scene.
For anchors (\cref{fig:camera_sampling}), we sample cameras on a cylinder looking at random directions.
CAT3D \cite{gao2024cat3d} is not open-sourced so we use our own baseline.

\inlineheading{Results}
We show qualitative results in \cref{fig:completing_casual_captures}.
We find that \textit{3DGS} cannot add any additional detail, leading to large unknown regions when rendering novel views away from the training images.
Naïvely adapting \textit{NeRFiller} to this challenging setting fails drastically because the inpainting is not conditioned on pose, so it is random and adds incorrect colors to the scene.
\textit{CAT3D-sequence-size} is more consistent but introduces artifacts due to the limited context size.
Our proposed method \textit{Fillerbuster}, with large sequence sizes for conditioning and generation, is the most consistent.
We design a metric for this task without ground truth, inspired by reconstruction error \cite{fridman2024scenescape} and epipolar geometry \cite{muller2024multidiff,yu2023long}.
We render a novel-view trajectory, estimate relative rotations between nearby frames using correspondences \cite{sun2021loftr} and classical methods \cite{hartley1997defense}, and report relative-rotation-accuracy (RRA): the percentage of pairs whose angle error is below a threshold.
RRA captures geometric stability but does not measure perceptual realism.

\begin{figure}[t]
\centering
\includegraphics[width=\linewidth]{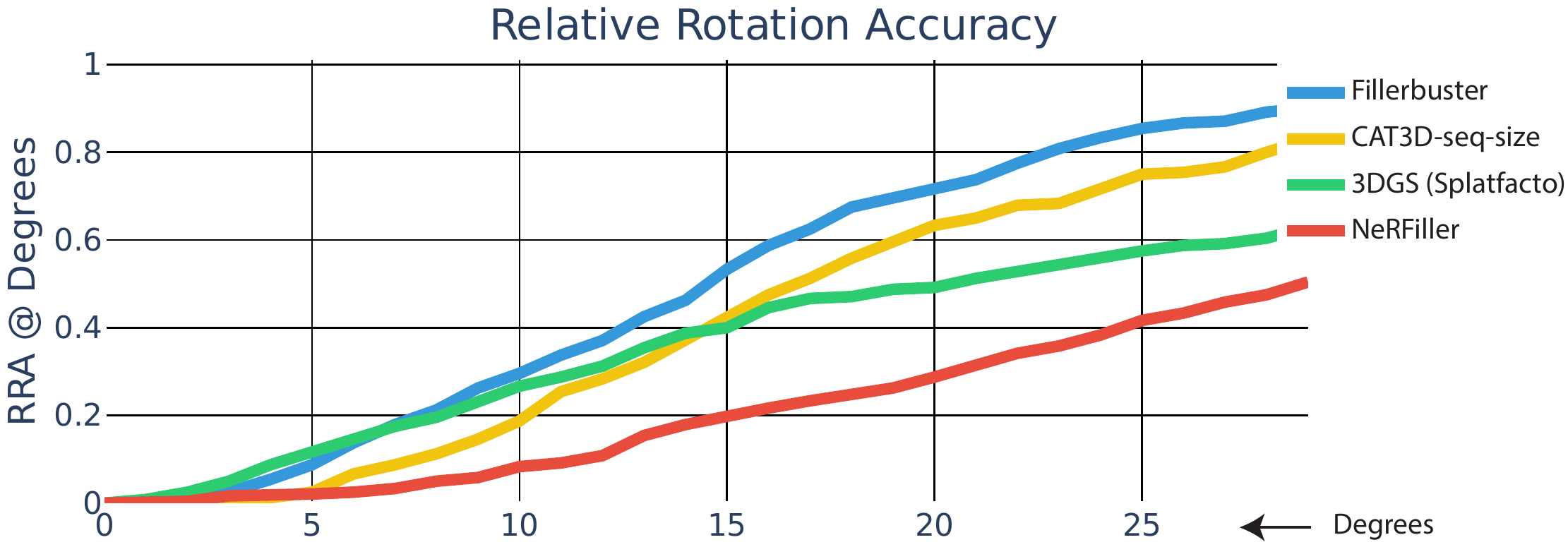}\vspace{-0.5em}
\caption{\label{fig:completing_casual_captures_metrics}%
    \textbf{Completing casual captures metrics.}
    We report relative rotation accuracy for nearby frames in a novel-view video.
    We use off-the-shelf correspondences \cite{sun2021loftr}
    to estimate camera rotation and compare with the ground truth.
    \method{} produces the most consistent videos from a pose-estimation perspective.
    \vspace{-2em}
}
\end{figure}

\subsection{Uncalibrated Scene Completion}

Here we consider a new task of ``uncalibrated scene completion'', starting from a collection of 16 unposed photos.
Our unified image-and-pose model supports such casual captures by predicting camera poses and then generating a fly-through of the scene, completing unknown content where missing.

\begin{figure*}[t]
\centering
\includegraphics[width=\linewidth]{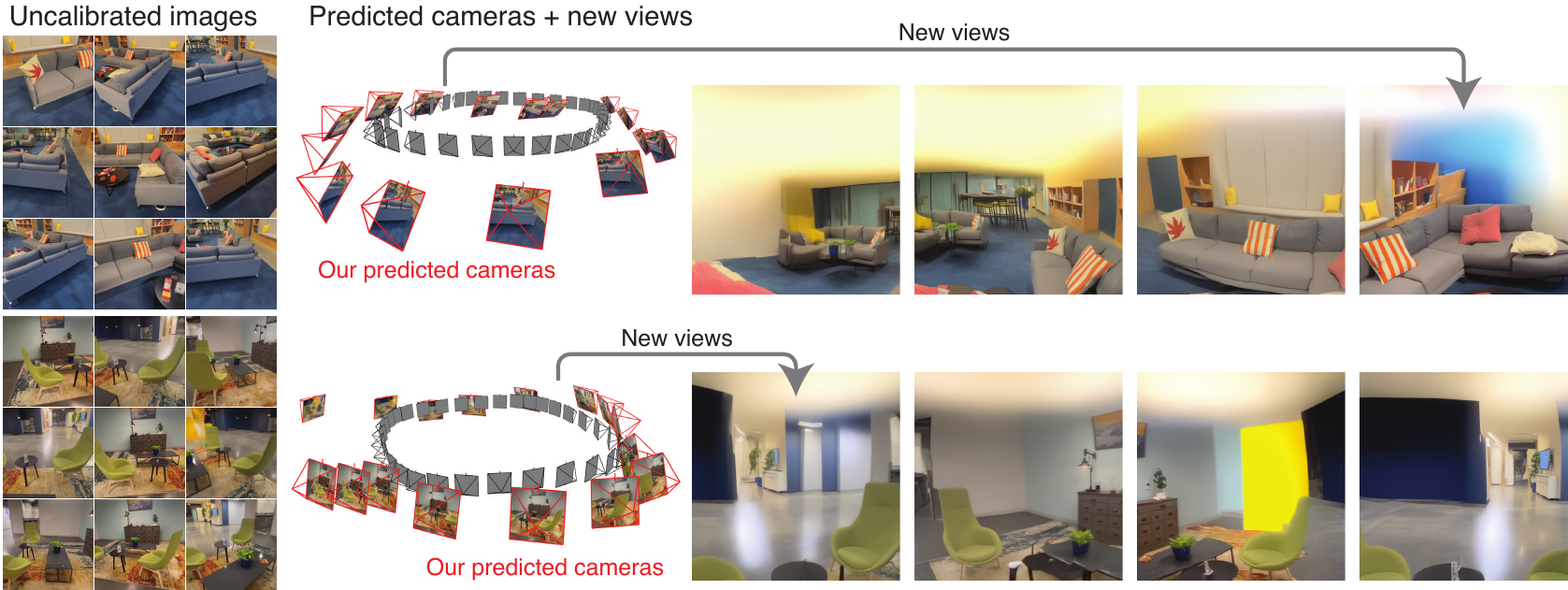}
\vspace{-15px}
\caption{\label{fig:uncalibrated_scene_completition}%
    \textbf{Uncalibrated scene completion.}
    We capture some scenes with an iPhone 14 Pro and run our framework. We start from 16 uncalibrated and unposed images (left), and we use our model to both predict camera pose (middle) and generate completed views (right).
    We show our predicted cameras in red compared to unknown views we will sample in black.
    Our cameras are plausible and useful for conditioning on to generate new views.
    We show just 4 views here and the full videos on the project page.
}
\vspace{-15px}
\end{figure*}

\inlineheading{Setting}
Given the set of images, we can denoise the raymaps conditioned only on the images.
We use joint denoising tiling with a window size of 8 images and average 8 times per denoising step (see appendix).
Then, we solve for the pinhole camera parameters that match backprojected rays to the denoised rays, taking only 5 seconds to converge for 16 images.
Next, we condition on our predicted rays to generate novel views to complete the scenes.
We create a camera path by fitting a 2D ellipse to our posed images and point cameras inward.
Here we use a sequence size of 48: 16 input images with generated poses, plus 32 generated images with specified poses, but note that this decision is flexible.

\begin{figure}[t]
\centering
\includegraphics[width=\linewidth]{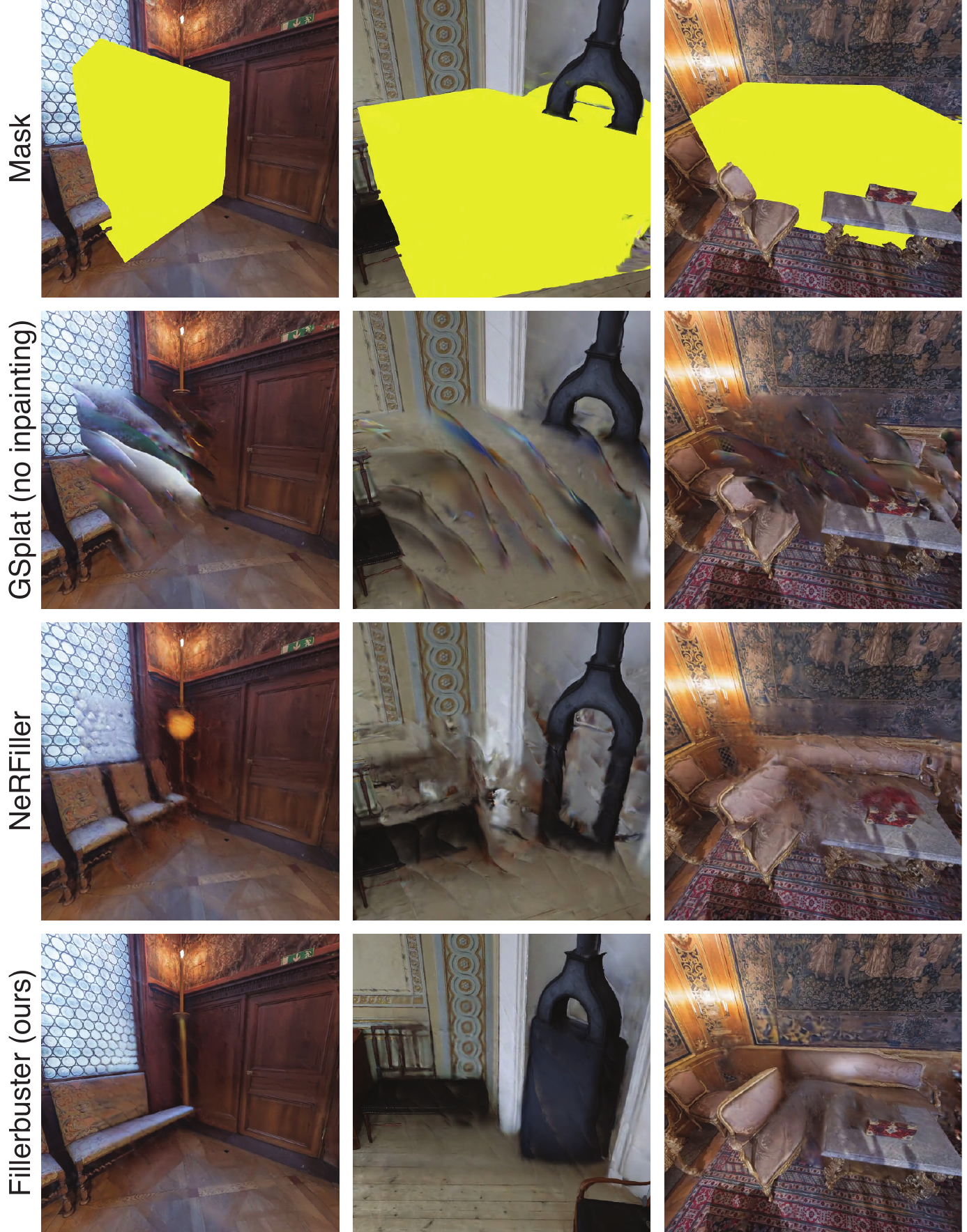}
\vspace{-15px}
\caption{\label{fig:nerfiller_results}%
    \textbf{NeRFiller dataset novel-views.}
    We complete NeRFiller scenes \cite{weber2024nerfiller} with higher quality and control than their method.
}
\vspace{-2em}
\end{figure}

\inlineheading{Results}
Our joint modeling of poses and images is convenient because we do not rely on external structure-from-motion; instead, our unified model can handle both tasks gracefully.
We show qualitative results of our video poses in \cref{fig:uncalibrated_scene_completition}.
Please see our project page for video results. We do not compare with COLMAP or SfM methods because they do not complete scenes, meaning they don't infill the missing areas outside the provided input images context.

\begin{table}[t]
\caption{\label{tab:completing_regions}%
    \textbf{Completing masked 3D regions.} 
    On the NeRFiller dataset \cite{weber2024nerfiller}, we report novel-view synthesis metrics where we compare the rendered images with the inpainted images.
    In parentheses, we report numbers without using our normal regularizations.
    No normal regularization lets the network cheat to explain inconsistencies, leading to slightly improved but misleading metrics.
    Overall, we find \method is much more consistent than NeRFiller.
}
\centering
\small
\begin{tabular}{lrrr}
\toprule
Method              & PSNR $\uparrow$ & SSIM $\uparrow$ & LPIPS $\downarrow$ \\ \midrule
NeRFiller             & 25.57 (25.94) & 0.89 (0.88) & 0.182 (0.194) \\
\method                & \textbf{29.60} (30.65) & \textbf{0.92} (0.93) & \textbf{0.096} (0.069) \\
\bottomrule
\end{tabular}
\vspace{-1em}
\end{table}

\subsection{Completing Masked 3D Regions}

\Cref{fig:nerfiller_results} shows results where we inpaint scenes from the NeRFiller dataset and compare against the NeRFiller method \cite{weber2024nerfiller}.
For both NeRFiller and \method, we inpaint 32 equally-spaced training images and then train 3D Gaussian splatting with our normal regularization.
Notably, unlike in NeRFiller, we do not use depth supervision or iterative dataset updates.
We instead inpaint once at the start of training to directly assess multi-view inpainting quality, regardless of any SDS-style optimizations that encourage consistency.
We also report reconstruction metrics in \cref{tab:completing_regions} by comparing our 32 inpainted images with the final renderings from the same 32 viewpoints.
Our method is more consistent than NeRFiller, with and without our normal regularization.

\begin{table*}
\caption{\label{tab:model_design}%
	\textbf{Model design ablations.}
	We evaluate our model on posed images from the Nerfstudio Dataset \cite{tancik2023nerfstudio}.
	We show a \method prediction above the table, where we compare the generation vs. the ground truth for reconstruction metrics (PSNR/SSIM/LPIPS).
	For hallucination metrics, we report the conditional validation loss (VAL) as done by \citet{esser2024scaling}.
	Notably, we focus on image generation rather than pose prediction but find that not predicting pose (``no-pose-pred'') leads to worse results.
	See \cref{sec:model-design} for detailed descriptions.
}
\centering
\scriptsize
\begin{tabular}{lcllll|llll|llll}
\toprule
\multirow{2}{*}{\textbf{Method}} & \multirow{2}{*}{\textbf{Iters}} & \multicolumn{4}{c}{\bfseries 8-views} & \multicolumn{4}{c}{\bfseries 16-views} & \multicolumn{4}{c}{\bfseries 32-views} \\

 & & PSNR $\uparrow$ & SSIM $\uparrow$ & LPIPS $\downarrow$ & VAL $\downarrow$ & PSNR $\uparrow$ & SSIM $\uparrow$ & LPIPS $\downarrow$ & VAL $\downarrow$ & PSNR $\uparrow$ & SSIM $\uparrow$ & LPIPS $\downarrow$ & VAL $\downarrow$ \\
\midrule

 no-index-emb & 100K & 11.10 & 0.431 & 0.456 & 0.2394 & 14.06 & 0.450 & 0.400 & 0.2417 & 12.38 & 0.467 & 0.422 & 0.2438\\ 
 fixed-index-emb & 100K &  10.46 & 0.353 & 0.520 & 0.2517 & 12.14 & 0.390 & 0.491 & 0.2546 & 12.52 & 0.416 & 0.448 & 0.2556\\ 
 no-poses & 100K &  11.63 & 0.413 & 0.426 & 0.2386 & 14.73 & 0.461 & 0.384 & 0.2411 & 13.39 & 0.476 & 0.389 & 0.2431\\ 
 random-poses & 100K &  11.82 & 0.431 & \textbf{0.415} & 0.2384 & \textbf{16.18} & \textbf{0.487} & 0.333 & 0.2409 & \textbf{14.27} & 0.483 & 0.367 & 0.2430\\ 
 \method & 100K & \textbf{11.97} & \textbf{0.435} & \textbf{0.415} & \textbf{0.2383} & 15.81 & 0.481 & \textbf{0.329} & \textbf{0.2407} & 14.21 & \textbf{0.486} & \textbf{0.366} & \textbf{0.2426}\\
 \method & 1M &  12.77 & 0.442 & 0.381 & 0.2365 & 17.20 & 0.485 & 0.281 & 0.2388 & 14.13 & 0.498 & 0.352 & 0.2396\\ 

\bottomrule
\end{tabular}
\vspace{-10px}
\end{table*}

\vspace{-5px}
\subsection{Model Design Ablations}
\label{sec:model-design}

We evaluate with the Nerfstudio dataset \cite{tancik2023nerfstudio}.
In \cref{tab:model_design}, we report novel-view synthesis metrics for 256$\times$256 resolution images for varying sequence lengths.
For each scene, we randomly sample a sequence of size $N$ image crops of this resolution.
We condition on $N/4$ full crops and $N/4$ partial crops, and generate all the missing information (see \cref{fig:model_samples} for examples).
We repeat this procedure 50 times for sequences of length $N \!\in\! \{8,16,32\}$, and report averaged metrics.
Each model is trained from scratch for 100K iterations with sequences of $N \!=\! 8$ images.
Inspired by \citet{esser2024scaling}, we also report validation losses (``VAL'') on these samples for 20 equally-spaced time steps.
This metric captures generation quality unlike the other metrics that evaluate reconstruction.

\begin{figure}
\centering
\includegraphics[width=\linewidth]{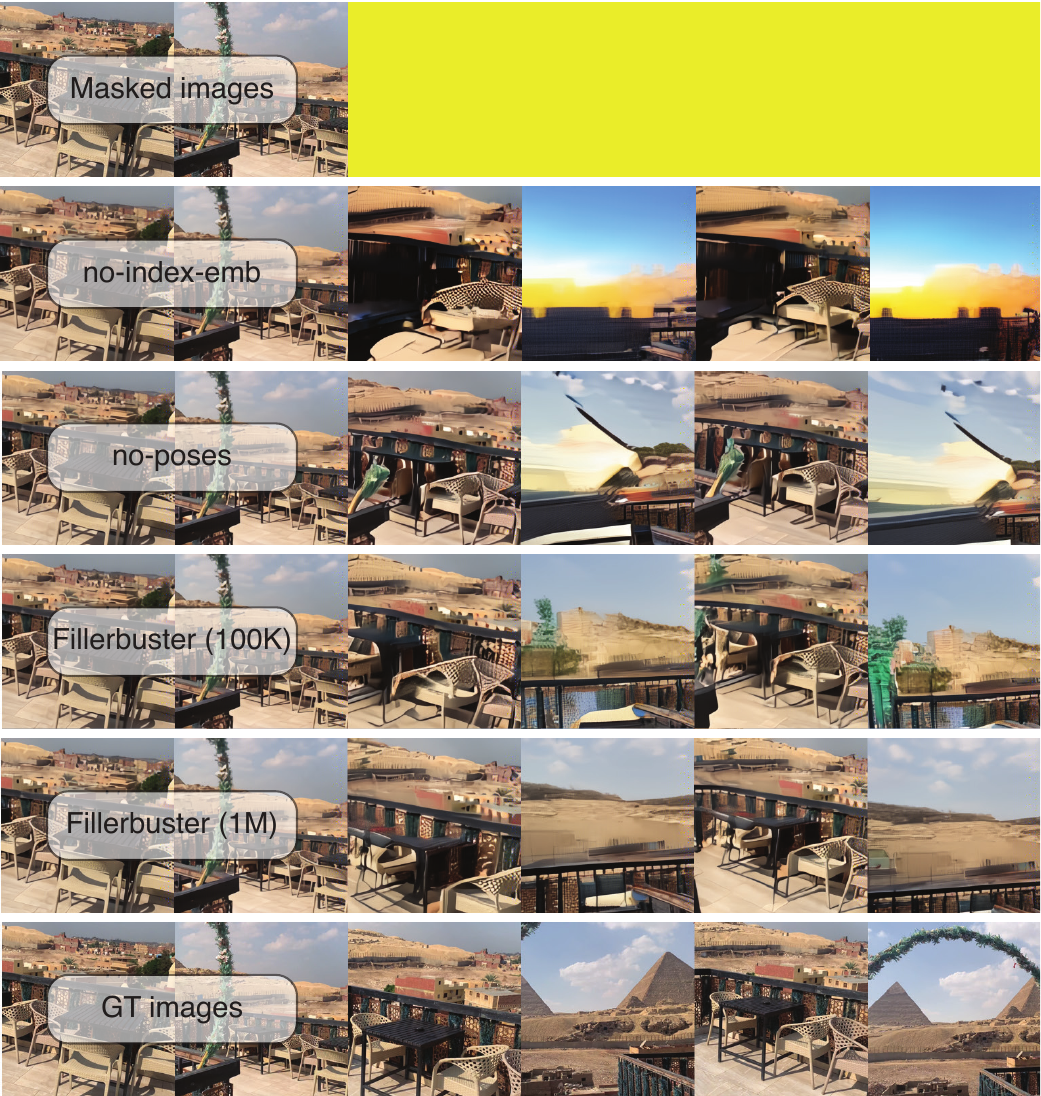}
\caption{\label{fig:qualitative_model_ablations}%
    \textbf{Qualitative results for model ablations}.
    We provide pose for all images and perform completion in the unknown regions (yellow).
    Without index embeddings, the model fails to reason about which image the tokens are coming from, so the results are patchy and blurry.
    Without training for pose prediction, the generations are worse than training for pose prediction (\method (100K)).
    The last rows show our final model and GT for reference.
}
\vspace{-20px}
\end{figure}

\Cref{tab:model_design} compares the following ablations (see \cref{fig:qualitative_model_ablations} for visual examples):
\textit{``no-index-emb"} does not use index embeddings and instead relies on the raymaps to understand the token relationships.
This makes the task harder and the model performs worse.
\textit{``fixed-index-emb"} uses a fixed number of index embeddings, preventing it from generalizing to more than 8 images.
To go beyond 8 images, more index embeddings are introduced, which this model variant cannot handle.
Notice the low PSNR of this setting for 16-views.
\textit{``no-poses"} does not denoise raymaps, and interestingly, we find that when it does not learn to predict camera pose, the model performs worse at image generation, indicating that image and pose predictions are complementary tasks.
Finally, \textit{``random-poses"} randomizes the poses instead of forcing them to be upright with one camera at the origin.
Our final model is trained for much longer and is shown at the bottom of the table, obtaining the best ``VAL'' results.
We also note that the metrics vary for 8, 16, and 32 views because each section in the table has different image subsets, making them non-comparable. The amount of known and unknown info also changes based on number of views, as described earlier.

\vspace{-10px}
\section{Conclusion}

Many 3D casual captures miss scene content because the camera does not look everywhere.
We presented \method{}, a large-scale multi-view diffusion model that completes missing regions and can recover camera poses when unavailable.
We showed results on casual capture completion, introduced the task of uncalibrated scene completion, and outperformed NeRFiller on masked 3D region completion.
Looking ahead, choosing camera paths is challenging because cameras should not be sampled inside objects or behind walls.
Our generations also degrade when sampled far outside the conditioning views, suggesting benefits from more diverse training data and stronger priors.
A method that predicts where to sample next could be valuable.
Rendering distant viewpoints in simulations may also help improve robustness.
We view this work as an initial step towards casual capture scene completion and expect quality to further improve with larger models, bigger compute budgets, and more diverse training data.
We trained our diffusion model from scratch due to legal constraints, but we expect leveraging pre-trained image/video diffusion models would lead to higher quality results when feasible.
We open-source our model weights, training, and processing code.

\vspace{0em}
\inlineheading{Acknowledgements}
We thank
Timur Bagautdinov,
Jin Kyu Kim,
Julieta Martinez,
Su Zhaoen,
Rawal Khirodkar,
Nir Sopher,
Nicholas Dahm,
Alexander Richard,
Bob Hansen,
Stanislav Pidhorskyi,
Tomas Simon,
David McAllister,
Justin Kerr,
Frederik Warburg,
Riley Peterlinz,
Evonne Ng,
Aleksander Holynski,
Artem Sevastopolsky,
Tobias Kirschstein,
Chen Guo,
Nikhil Keetha,
Ayush Tewari,
Changil Kim,
Lorenzo Porzi,
Corinne Stucker,
Katja Schwarz, and Julian Straub for helpful discussions and support.

{
    \small
    \bibliographystyle{ieeenat_fullname}
    \bibliography{main,Fillerbuster-CR}
}

\clearpage
\appendix
\setcounter{figure}{0}
\renewcommand{\thefigure}{A.\arabic{figure}}
\setcounter{table}{0}
\renewcommand{\thetable}{A.\arabic{table}}
\section{Appendix}
\label{sec:appendix}

Additional details are provided below. See also our project page at \url{https://ethanweber.me/fillerbuster/} for interactive results and videos.

\section{More Results}

Please see the website for interactive results where we show more results in the Nerfbusters and Nerfiller datasets. We also have a new experiment in a section called \textbf{``Flexible Conditioning and Generation"} where we illustrate how our model can adapt to different number of input frames. We show results for this on the LERF dataset~\cite{kerr2023lerf}.

\begin{figure*}[t]
\centering
\includegraphics[width=\linewidth]{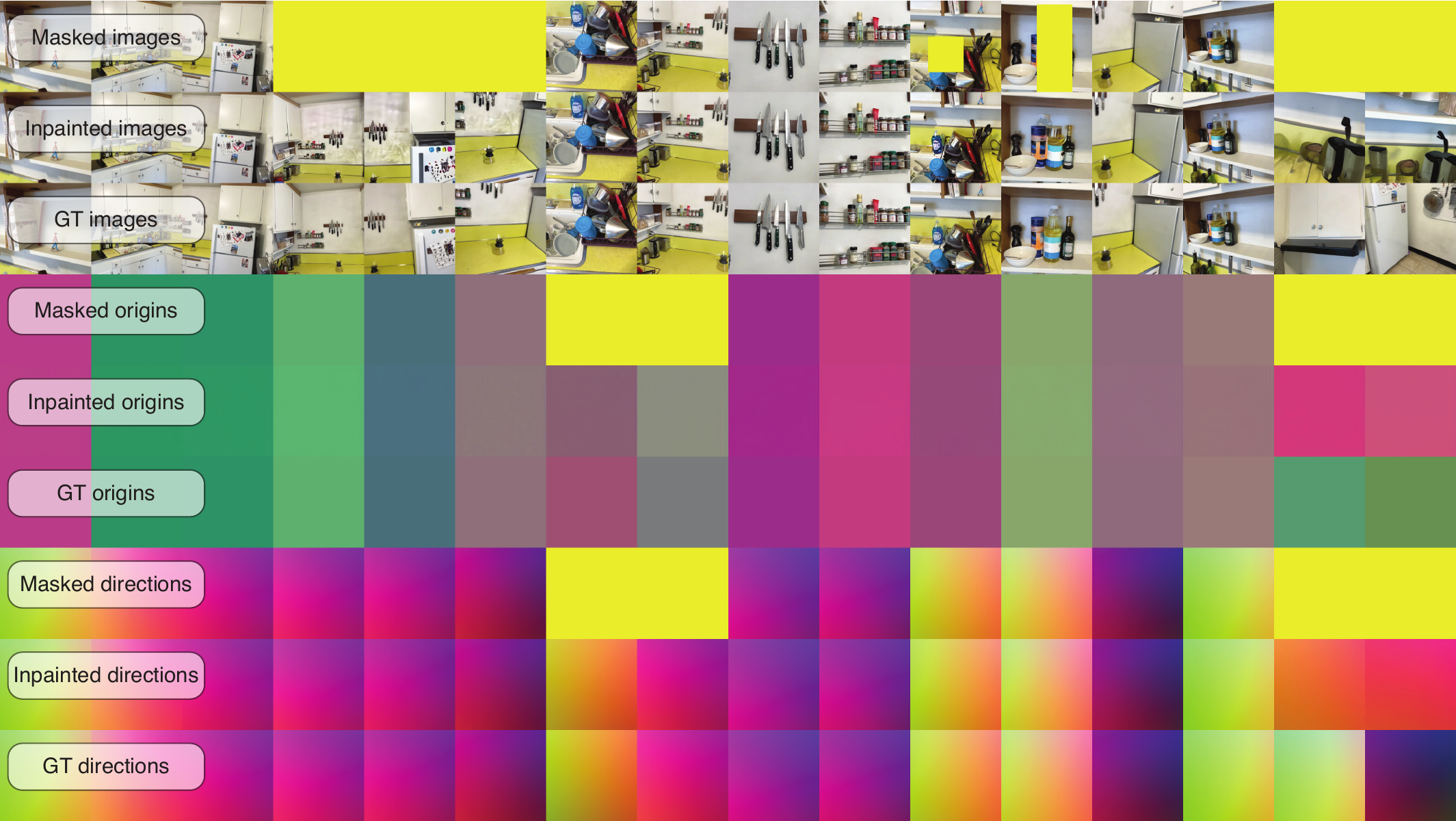}\vspace{-0.5em}
\caption{\label{fig:model_sample}%
    \textbf{Recovering missing information from a multi-view sequence.}
    Our model is flexible in that it can condition on any available information and recover the missing regions.
    We mark the conditioning as ``Masked images'', ``Masked origins'', and ``Masked directions''.
    The yellow regions are where information is not known.
    Given these conditions, we can recover all the missing information in the ``Inpainted'' rows.
    Notice that we are recovering fully unknown images, unknown poses, partial images, and we are generating two fully new images and poses at the same time (far right).
    The ground truth ``GT'' rows are provided as reference from the original capture, but the model only needs to follow the GT when the input provides the appropriate information. Our model does not require a ``GT'' reference and instead the model can be used to complete casually captured scenes where there is no reference, as shown in our paper.
    The origins and directions are all within the cube $[-1, 1]^3$ so can be visualized in RGB space.
}\vspace{-1em}
\end{figure*}

\section{Training Recipe}

Here we provide more training details about our model. Some of the data we use are as follows:
\begin{itemize}
  \item \textbf{ScanNet++~\cite{yeshwanth2023scannet++}:}
  $\sim$500 captures. This dataset is a collection of indoor scenes. We use their high-quality DSLR images data for training our model.

  \item \textbf{Shutterstock3D:}
  $\sim$2M 3D assets from Shutterstock. For the meshes, we render them from 24 views sampled on a sphere using Blender and the Cycles physically-based path tracer, similar to Objaverse \cite{deitke2022objaverse,deitke2023n}.

  \item \textbf{Shutterstock2D:}
  $\sim$400M image-text pairs from Shutterstock.
  These images are aesthetically pleasing. We use these images since they have text information and can help learn the long-tail of information that may not be present in our other datasets.
\end{itemize}

\noindent
We combine these datasets with various other multi-view data available to us. We train our model for multi-view image and raymap denoising.
We include single-view data in the training mix because we are training the model from scratch and wish to learn more concepts outside of of the more specific multi-view data.
50\% of the time, the model samples multi-view data, and the other 50\% of the time, the model samples single-view data.

\inlineheading{Multi-view data sampling}
We implement our multi-view dataloader by leveraging the Nerfstudio \cite{tancik2023nerfstudio} framework.
We use custom Nerfstudio DataParsers for each dataset type and train NeRFs on a subset of the captures with the Nerfacto \cite{tancik2023nerfstudio} method.
After confirming that our image crops and rays are sampled properly, we can confidently use the dataset for training our multi-view diffusion model.
Within each of these multi-view datasets, we uniformly sample frames with one of 5 strides: 1, 2, 4, 8, or completely random (i.e. no stride).
We sample random crops within each image, and with 10\% chance we center the crop.
With 10\% chance, we drop out conditioning.
75\% of the time, we train for image denoising, and 25\% of the time, we train for raymap denoising.

\inlineheading{Single-view data sampling}
Our single-view data from Shutterstock2D is treated as a single-sequence set with a text prompt and unknown camera ray conditioning.
In this case, we mask out the raymaps as conditioning and we also mask out the loss for the noise prediction, to not penalize the predictions where we do not have ground truth.
We use cross-attention with text embeddings.
In practice, we only use the text to control coarse signals like like brightening the generation by using the word ``bright".

\inlineheading{Training}
We train with mixed-precision, using \texttt{bfloat16} to reduce memory requirements and speed up training.
To stabilize training, we found it was important to perform LayerNorm in \texttt{float32}, and to normalize keys and queries before attention operations~\cite{esser2024scaling}.
We use multi-node DDP on a Slurm cluster with GPU-based dataloading.

\inlineheading{Hyperparameters}
We use flow matching and logit normal sampling \cite{esser2024scaling}.
Our learning rate is constant at $10^{-4}$.
For multi-view training at 256\textsuperscript{2} resolution, we use a batch size of 4 sequences, and a sequence size of 10 for each GPU.
For single-view training at 256\textsuperscript{2} resolution, we use a batch size of 52 images, where in this case the sequence size is 1.
We train with 64 A100 GPUs across 8 nodes for 1M steps which takes roughly a month.
Finally, we train for an additional 100K steps with varying image resolution.
To implement this, we assign each GPU to a specific resolution with uniform probability and tune the batch size and sequence size to fit within the GPU memory. We use the sequence sizes \{20, 20, 10, 5, 2\} for resolutions \{64\textsuperscript{2}, 128\textsuperscript{2}, 256\textsuperscript{2}, 512\textsuperscript{2}, 1024\textsuperscript{2}\}, respectively.

\noindent\textbf{Loss and guidance weights.}
Image/raymap VAEs use KL, adversarial, and L1 terms with $\lambda_\text{KL}{=}10^{-6}$, $\lambda_\text{adv}{=}0.5$, and $\lambda_\text{L1}{=}1$.
Classifier-free guidance dropout is 10\%, equally likely to remove images, raymaps, or both.
At inference, we use spatially varying CFG weights of 7 for unknown regions and 1.1 for known regions.
For normal regularization in 3DGS training, we set $\lambda_\text{align}{=}0.01$.

\section{Method Details}

\subsection{Image and Raymap VAE}

We train two separate VAEs from scratch for compressing images and raymaps into latent representations.
We use the same convolutional architecture \cite{rombach2022high} for both VAEs, each with 4 down blocks and 4 up blocks.
For the image VAE, the output channels for each down block are 128, 256, 512, and 512.
We use 3 input channels and 16 dimensions for the latent dimension.
For the raymap VAE, we reduce the down block channel dimensions by a factor of 4, leading to dimensions: 32, 64, 128, and 128.
The raymap VAE takes as input 6 channels and has a latent dimension of 16.
For the raymap VAE, we remove group norm since we noticed it produced spot artifacts in the corner, consistent with previous findings \cite{MovieGen2024}, and we modified the padding in the network to use ``replicate'' padding.
The image VAE has 84M parameters and the raymap VAE has 5M parameters. The image VAE trains on all our data, while the raymap VAE trains on only our multi-view data.

\subsection{Model Architecture}

\begin{figure*}[t]
\includegraphics[width=\linewidth,trim=0 200 0 140,clip]{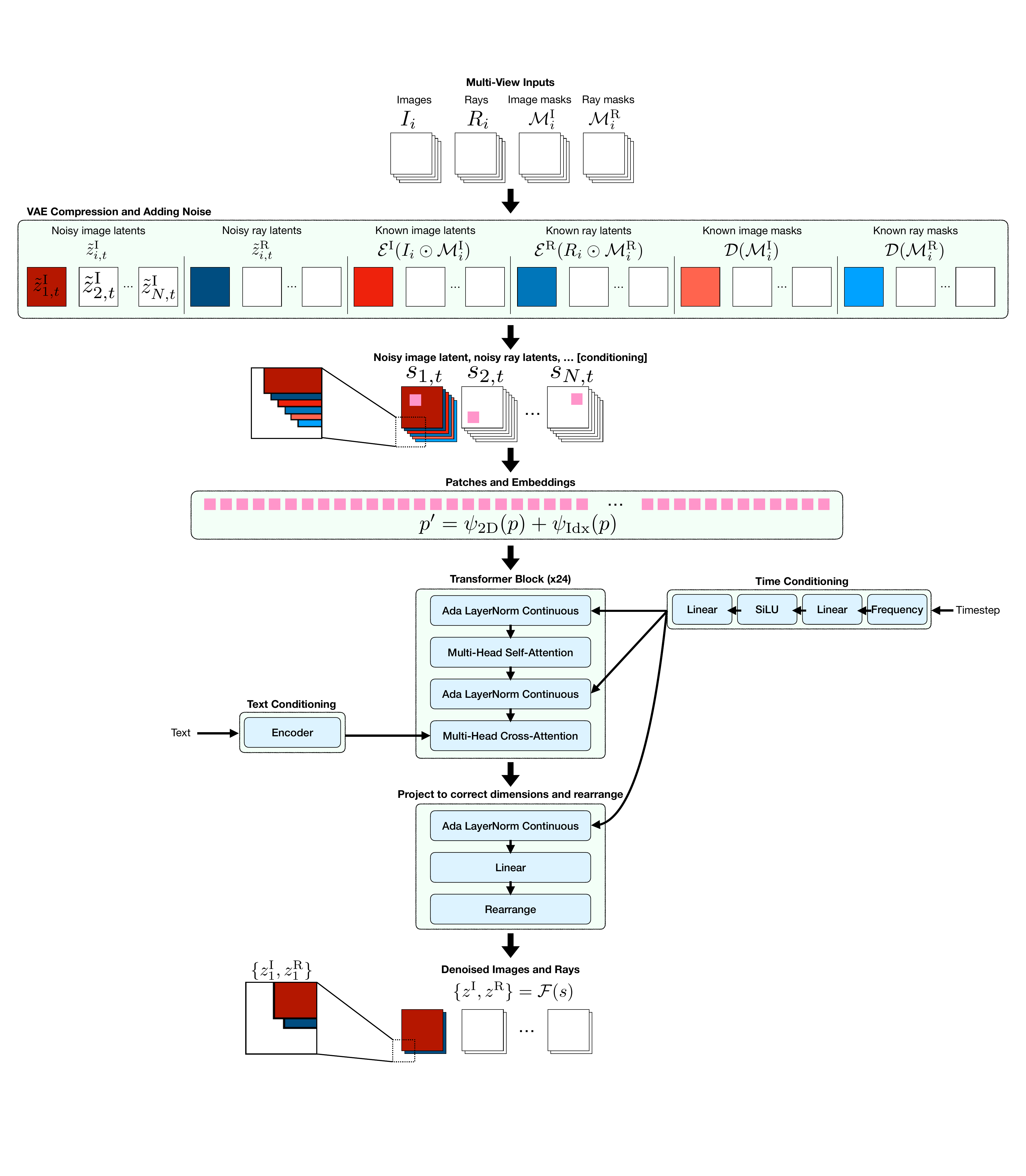}
\caption{\label{fig:model_architecture_all}%
    \textbf{\method{} model architecture.}
    Here we show the full model architecture from input (top) to output (bottom).
    We use the same notation as the main paper, where $I$ denotes images, $R$ denotes raymaps, and $\mathcal{M}$ denotes masks indicating where we know information or not.
    We inject various conditioning time and text conditioning into the model, as shown on the sides of the transformer.
    Timestep conditioning is necessary because our model is a diffusion transformer.
    Text, however, is optional.
    We include it to jointly train on our single-view image collection, which has text annotations, whereas the multi-view sequences are always trained with an empty string as the text prompt.
    The ``Ada LayerNorm Continuous'' is normalization with scale and shift modulation from the continuous time conditioning.
}
\vspace{8.2em}
\end{figure*}

We provide an overview figure for our model in the main paper.
In the appendix, we provide a more detailed figure of our model in \cref{fig:model_architecture_all}.
Please see the figure and its caption for more details.
We also provide a model sample in \cref{fig:model_sample} showing a combination of known images, missing images, missing poses, partial images, and where both images and poses are unknown, resulting in unconditional generations.

\subsection{Raymap Prediction Discussion}

We solve for camera poses with a MultiDiffusion-style approach \cite{muller2024multidiff,weber2024nerfiller}, where we use a smaller sequence size as input to the model and average predictions in order to predict a larger context size.
We found this approach to give higher quality results for raymap recovery compared to passing in all images together in the same forward-pass.
However, this finding is not true for generating image content.
For the image prediction task, it's actually better to pass in all images in the same forward-pass.
We suspect this finding is because relative camera pose estimation will be unaffected by cameras that do not look in the same areas, but image generation will still be influenced by all context in the scene.

\subsection{Inference Speed}

We provide inference speeds in \cref{tab:inference_speeds}.
We report this on a NVIDIA A100.
Here we sample for 50 steps, but in practice one could sample for 24 steps and achieve similar quality results, while halving the inference time. Our model is on the order of seconds to generate a handful of multi-view images, which makes it useful to use with open-sourced reconstruction frameworks~\cite{tancik2023nerfstudio}.

\begin{table}[H]
\caption{\label{tab:inference_speeds}%
    \textbf{Inference time of our model.} 
    Measured for 50 denoising steps, including time for VAE encoding and decoding.
}
\centering
\small
\setlength{\tabcolsep}{5pt}
\begin{tabular}{lrrrr}
\toprule
Resolution              & 4 Images & 8 Images & 16 Images & 32 Images \\ \midrule
256$\times$256          & 9 sec & 9 sec & 12 sec & 23 sec \\
512$\times$512          & 12 sec & 23 sec & 53 sec & 2 min \\
\bottomrule
\end{tabular}
\vspace{-1em}
\end{table}

\section{Evaluation Details}

\subsection{CAT3D-Sequence-Size Baseline}

For this baseline, we choose 3 images from a casual capture and use them for conditioning.
Then, we generate 20 rounds of inpaints, generating 6 new views each time to reach a total of 120 new generations, the same number that we generate with our method that uses larger sequence sizes.
The 6 target views are sampled at a random elevation and height on a cylinder.
In practice, CAT3D \cite{gao2024cat3d} uses 5 new views, but we use 6 to ease comparison in our implementation.

\subsection{Relative Rotation Accuracy Metric}

We use LoFTR~\cite{sun2021loftr} (outdoor weights) and Kornia~\cite{riba2020kornia} to recover the relative rotation between frames sampled within 1 second of each other (in our 10-second videos).
We sample 20 pairs per capture and report the average Relative Rotation Accuracy for our baselines.

\subsection{Nerfbusters Dataset}

We use the train splits of the Nerfbusters dataset \cite{weber2024nerfiller}.
Nerfbusters was captured with a training and evaluation video, but we only consider the training video, since it mimics the casual capture of an inexperienced photogrammetry user.

\subsection{NeRFiller Dataset}

We obtain the NeRFiller dataset and camera paths from the NeRFiller project \cite{weber2024nerfiller}.
We exclude the ``backpack'' capture for our analysis because it failed to reconstruct with Splatfacto~\cite{tancik2023nerfstudio,ye2024gsplat}. We suspect this is because the ``backpack'' capture is a forward-facing scene from SPIn-NeRF \cite{mirzaei2023spin}.

\section{Normal Regularization}

\begin{figure}[H]
\centering
\includegraphics[width=\linewidth]{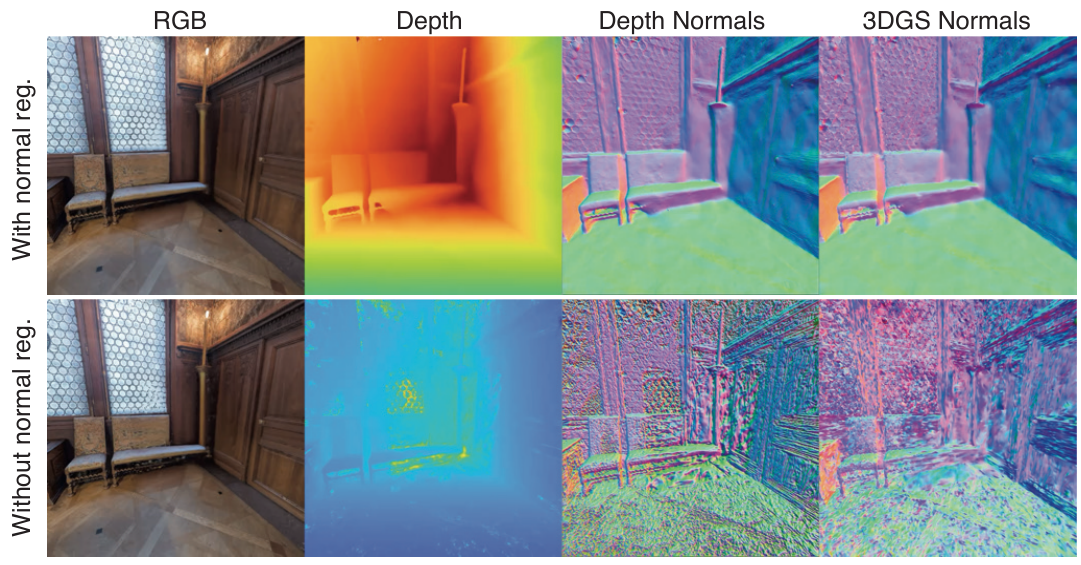}
\caption{\textbf{Normal regularization.} We find that aligning depth-derived normals with 3D Gaussian Splatting normals helps improve our reconstructions.}
\label{fig:normal_regularization}
\end{figure}

Our normal regularization loss $\mathcal{L}_\text{align} = \|\mathrm{sg}(N_\text{r}) - N_\text{d}\|^{2}_{2} + \|N_\text{r} - \mathrm{sg}(N_\text{d})\|^{2}_{2}$ aligns 3D Gaussian normals with depth-derived normals. Specifically, $N_\text{r}$ are rendered normals from 3D Gaussians, oriented towards the camera, and $N_\text{d}$ are normals derived from rendered depth maps by backprojecting rendered depth and computing cross products for nearby points. We found that applying this regularization at the start of training can prevent the Gaussians from moving around gracefully during the start of optimization, however, if we start the regularization after 10K steps when the initial structure settles, we can significantly improve geometry. \cref{fig:normal_regularization} shows our results with and without normal regularization.
Note that we also use a TV loss to help create smoother surfaces rather than jagged ones, but it has little effect on the geometry overall.

\end{document}